\documentclass{ecai}
\usepackage{graphicx}
\usepackage{latexsym}
\usepackage{amsmath}

\usepackage{amssymb} 
\usepackage{utfsym}
\usepackage{booktabs}
\usepackage{makecell}
\usepackage{multirow}
\usepackage{algorithm}
\usepackage{listings}
\usepackage{tikz}

\def\ie{\emph{i.e., }}
\def\eg{\emph{e.g., }}

\begin{document}

\begin{frontmatter}

\title{Identifying the Defective: Detecting Damaged Grains for Cereal Appearance Inspection}

\author[A,B,$\dagger$]{\fnms{Lei}~\snm{Fan}\thanks{Work done when interning at Gaozhe Technology. \\ Corresponding Author. Email: lei.fan1@unsw.edu.au} \orcid{0000-0001-9472-7152}}

\author[A,$\dagger$]{\fnms{Yiwen}~\snm{Ding}}
\author[A]{\fnms{Dongdong}~\snm{Fan}} 
\author[A]{\fnms{Yong}~\snm{Wu}}
\author[B]{\fnms{Maurice}~\snm{Pagnucco}}
\author[B]{\fnms{Yang}~\snm{Song}}

\address[A]{Gaozhe Technology, Hefei, China}
\address[B]{The School of Computer Science and Engineering, UNSW Sydney, Sydney, Australia}
\address[$\dagger$]{Equal Contribution.}

\begin{abstract}
Cereal grain plays a crucial role in the human diet as a major source of essential nutrients. Grain Appearance Inspection (GAI) serves as an essential process to determine grain quality and facilitate grain circulation and processing. However, GAI is routinely performed manually by inspectors with cumbersome procedures, which poses a significant bottleneck in smart agriculture.

In this paper, we endeavor to develop an automated GAI system: \textbf{AI4GrainInsp}. By analyzing the distinctive characteristics of grain kernels, we formulate GAI as a ubiquitous problem: Anomaly Detection (AD), in which healthy and edible kernels are considered \textit{normal} samples while damaged grains or unknown objects are regarded as \textit{anomalies}. We further propose an AD model, called AD-GAI, which is trained using only normal samples yet can identify anomalies during inference. Moreover, we customize a prototype device for data acquisition and create a large-scale dataset including $220K$ high-quality images of wheat and maize kernels. Through extensive experiments, AD-GAI achieves considerable performance in comparison with advanced AD methods, and \textbf{AI4GrainInsp} has highly consistent performance compared to human experts and excels at inspection efficiency over 20$\times$ speedup. The dataset, code and models will be released at \textbf{https://github.com/hellodfan/AI4GrainInsp}.

\end{abstract}
\end{frontmatter}

\section{Introduction}

\begin{figure}[!t]
	\centering
	\begin{center}
		\includegraphics[width=0.44\textwidth]{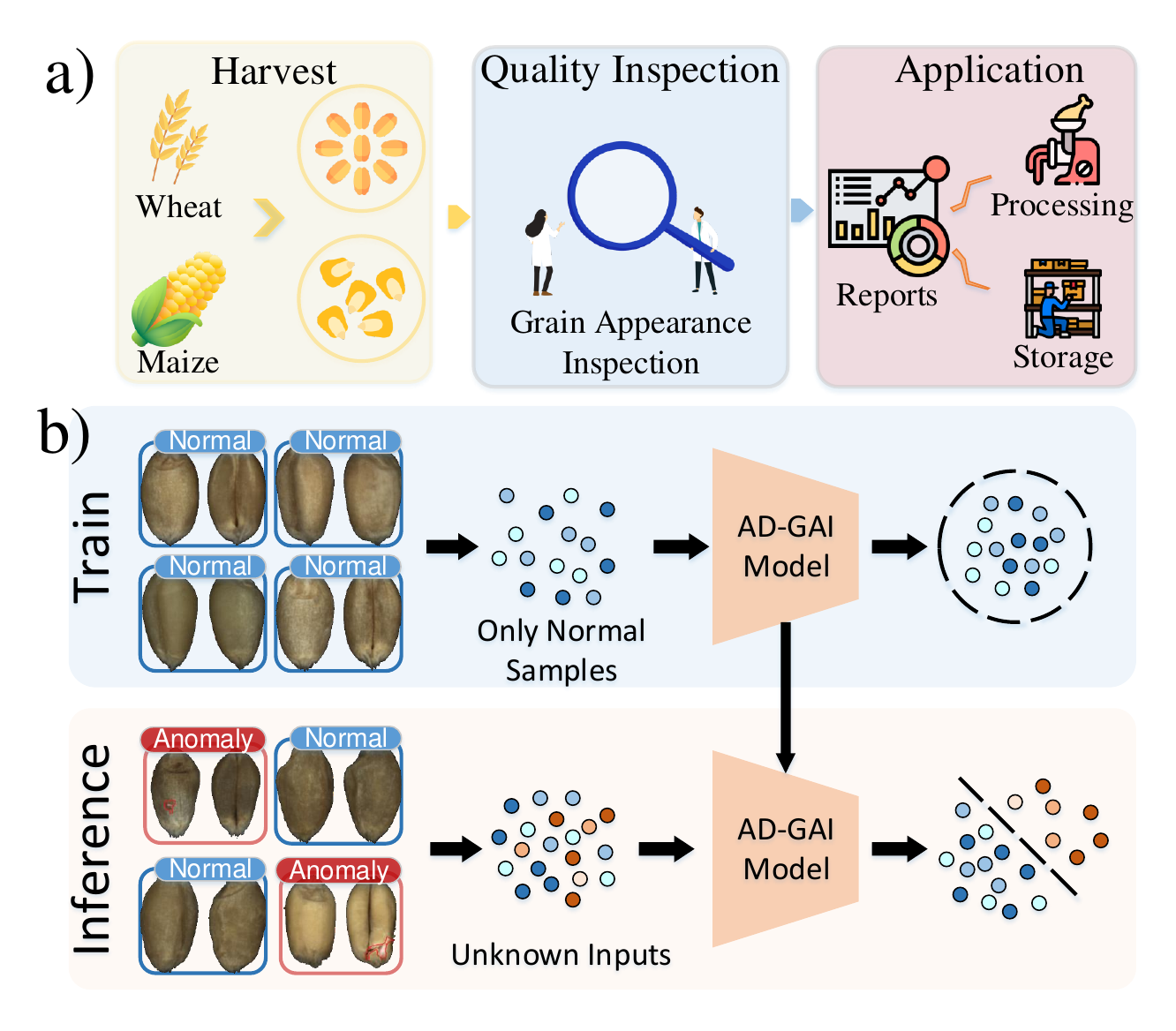}
	\end{center}
    \vspace{-.5em}
	\caption{a) Role of GAI in agriculture. b) GAI is formulated as an AD problem. Only \textit{normal} samples are used for training, while models require identifying whether the test samples are \textit{normal} or \textit{anomalous} samples.}
    \vspace{-.5em}
	\label{fig:grain_examples}
\end{figure}

Cereal grain plays a critical role in human survival and the development of civilizations, ensuring a reliable supply of food, contributing to poverty eradication and providing essential ingredients for various food products and daily necessities. The Quality Inspection of cereal Grains (QIG) is of paramount importance for standardizing grain storage, promoting fair circulation and guiding processing. It serves as a crucial metric for assessing nutrition, ensuring the security of supply, and identifying stratification (see Figure~\ref{fig:grain_examples}.a). 
Furthermore, QIG reflects crop conditions and holds the potential to guide sustainable and eco-friendly practices in smart agriculture. 
Currently, there are two dominant QIG methods: Chemical Analysis (CA) and Grain Appearance Inspection (GAI). CA is based on molecular biology and chemistry along with chemical substances and laboratory equipment, enabling highly precise inspection. On the other hand, GAI relies on visual characteristics to assess the appearance of grain kernels. Compared to CA, GAI is overwhelmingly adopted for high-throughput determination of the quality of cereal grains, including the detection of impurities, extraneous cereals, moldy grains and other damaged grains, as defined in the cereal ISO standard \cite{ISO5527}.

\begin{table*}[!t]
	\centering
    \vspace{-.5em}
	\caption{Wheat and maize examples of healthy, damaged grains and impurities (abbreviation used in the subsequent content).}
    \vspace{-.5em}
	\resizebox{0.98\textwidth}{!}{
		\begin{tabular}{cccc}\toprule
			\makecell[c]{\underline{H}ealth\underline{Y} grain (HY)} 	& \makecell[c]{\underline{S}proute\underline{D} grain (SD)} & \makecell[c]{\underline{F}usarium\&\underline{S}hriveled grain (F\&S)}   &
   \makecell[c]{\underline{B}lack \underline{P}oint (BP) grain for wheat \\ \underline{H}eate\underline{D} (HD) grain for maize }  \\ \cmidrule(lr){1-4}
			\begin{minipage}[b]{0.7\columnwidth}
				\centering
				\raisebox{-.42\height}{
					\includegraphics[width=\linewidth]{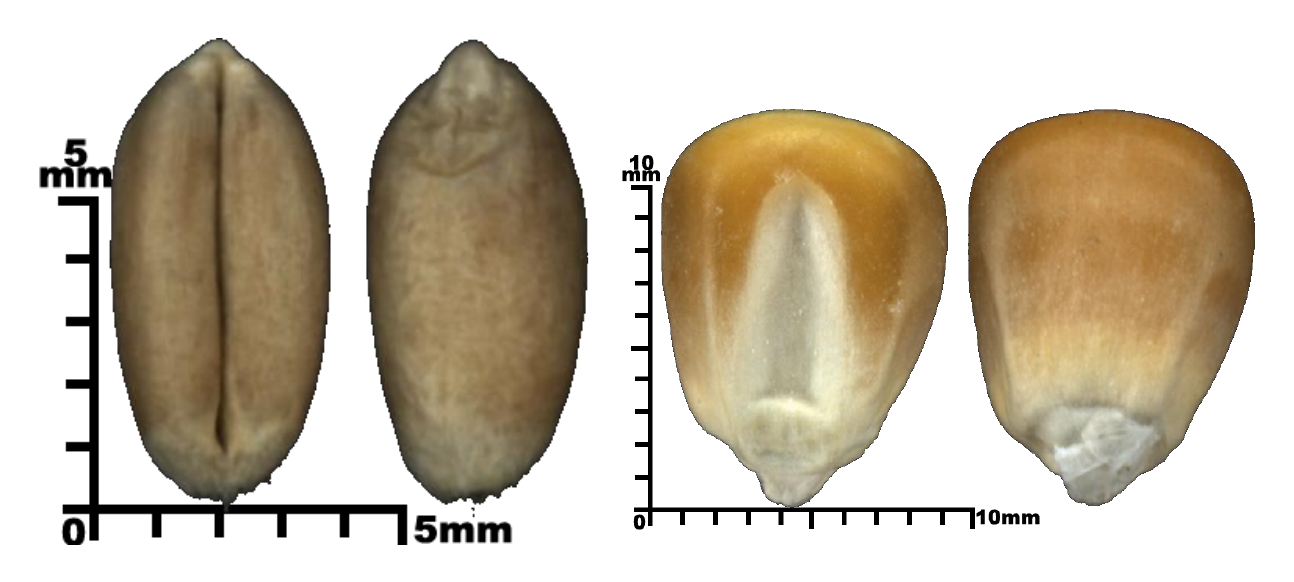}}
			\end{minipage} 
			& 
			\begin{minipage}[b]{0.7\columnwidth}
				\centering
				\raisebox{-.4\height}{\includegraphics[width=\linewidth]{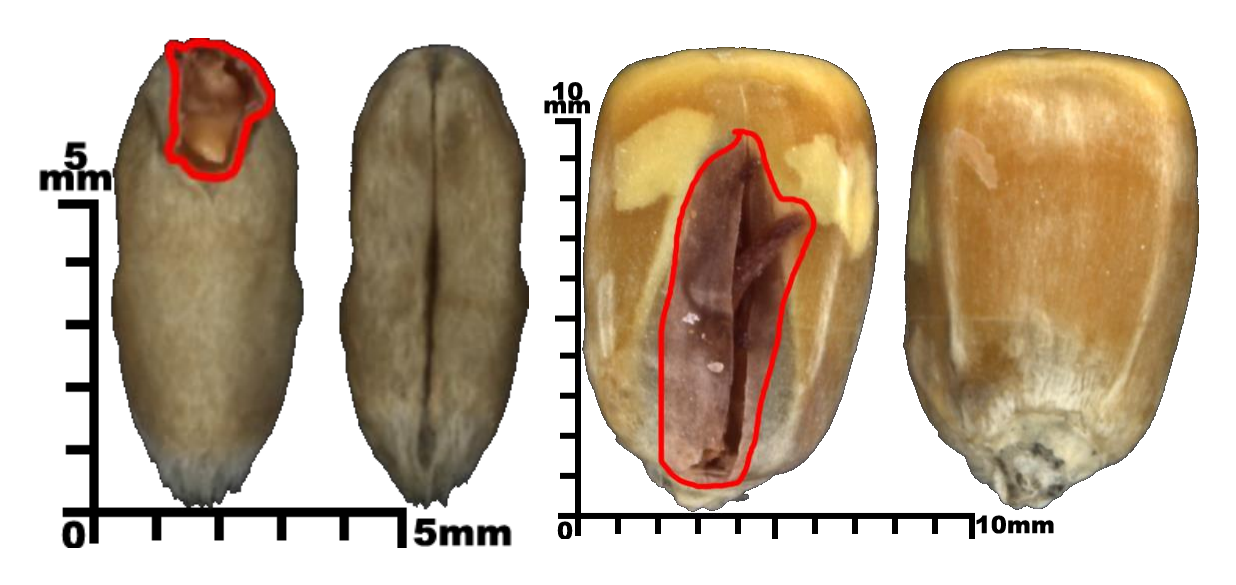}}
			\end{minipage}
			&
			\begin{minipage}[b]{0.7\columnwidth}
				\centering
				\raisebox{-.45\height}{\includegraphics[width=\linewidth]{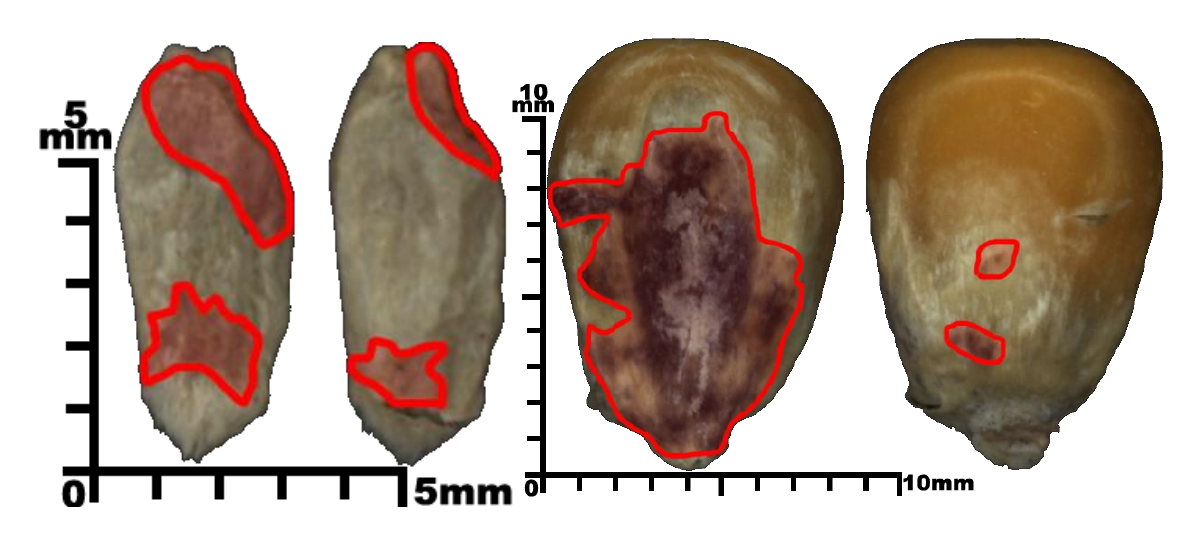}}
			\end{minipage}	
			&   
			\begin{minipage}[b]{0.7\columnwidth}
					\centering
					\raisebox{-.46\height}{
						\includegraphics[width=\linewidth]{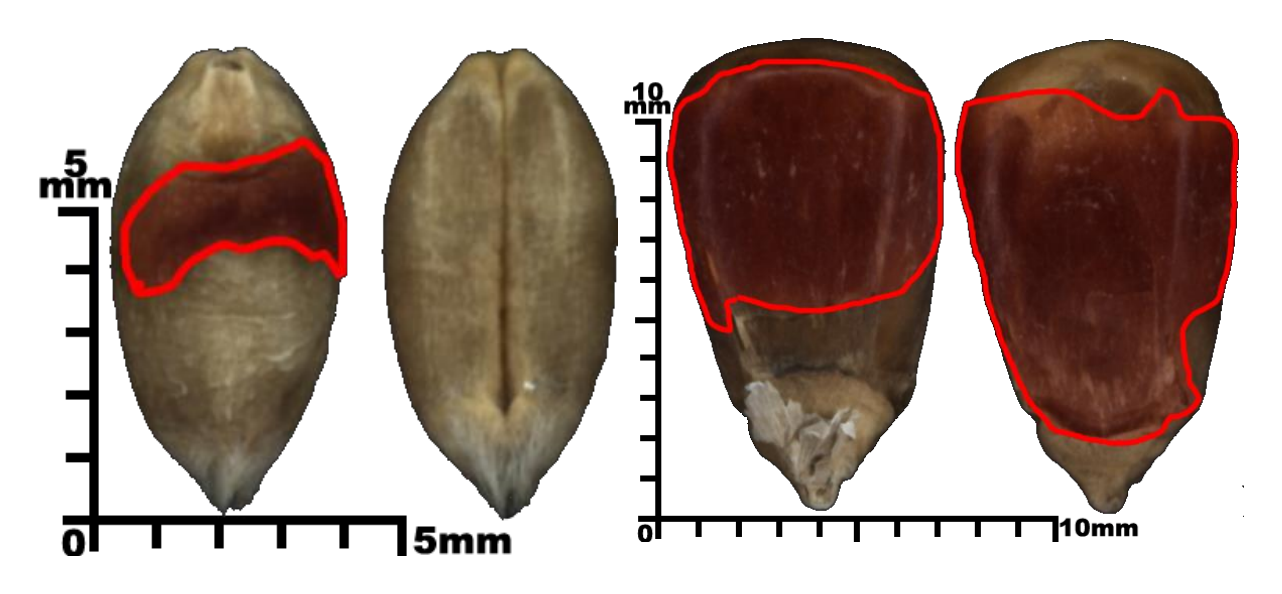}
					}
			\end{minipage}

			\\	\cmidrule(lr){1-4}
				 \makecell[c]{\underline{M}old\underline{Y} grain (MY)}  &  \makecell[c]{\underline{B}roke\underline{N} grain (BN)} &  \makecell[c]{Grain \underline{A}ttacked by \underline{P}ests (AP)} & \makecell[c]{\underline{IM}purities (IM)}  \\\cmidrule(lr){1-4}
 			\begin{minipage}[b]{0.7\columnwidth}
 				\centering
 				\raisebox{-.45\height}{\includegraphics[width=\linewidth]{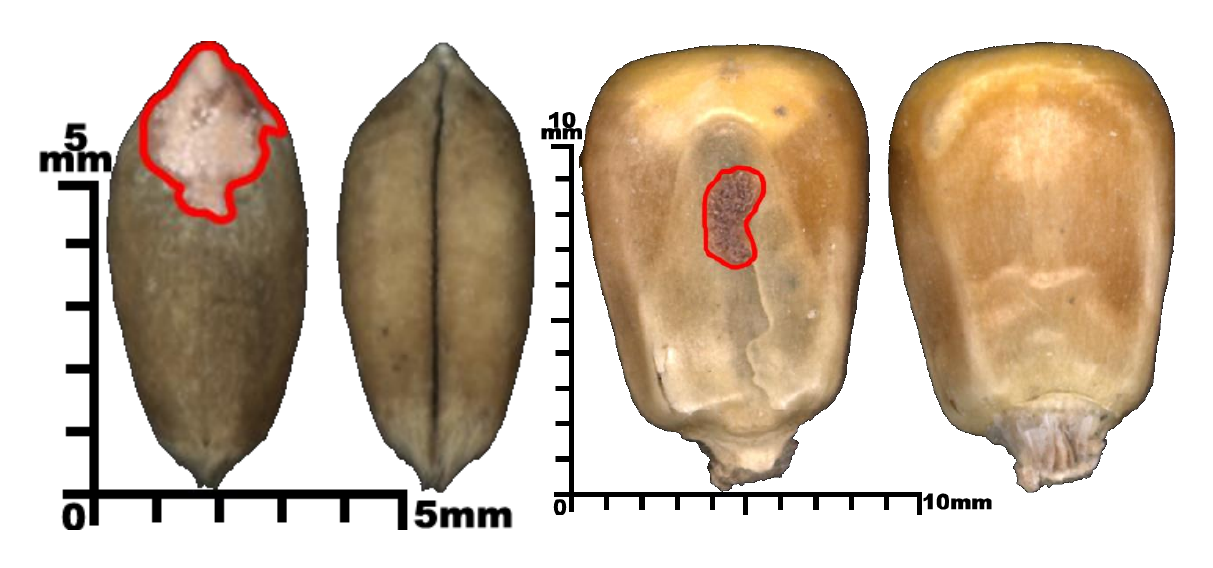}}
 			\end{minipage}  
			&  
			\begin{minipage}[b]{0.7\columnwidth}
				\centering
				\raisebox{-.51\height}{\includegraphics[width=\linewidth]{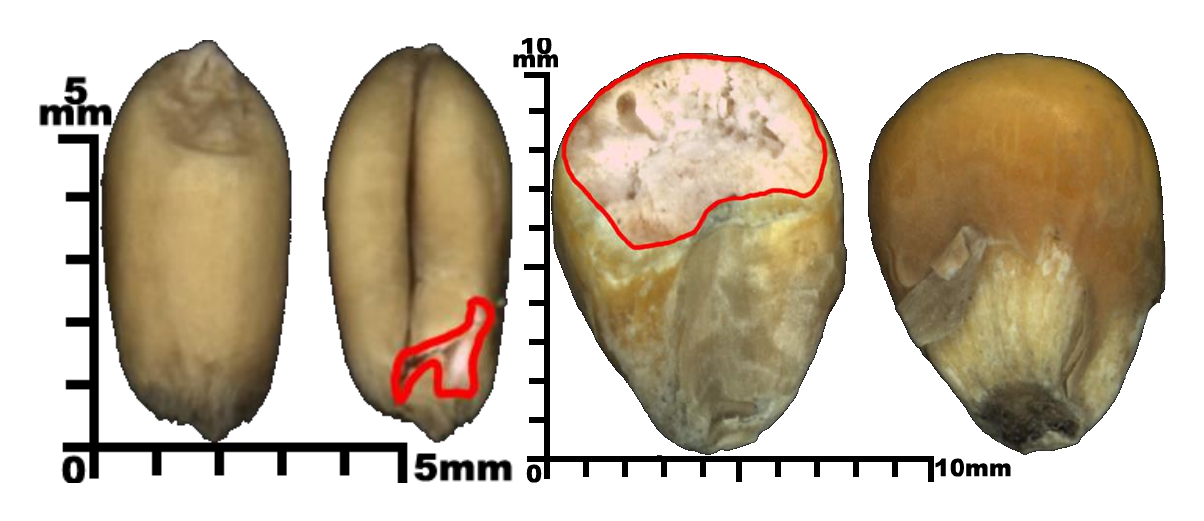}}
			\end{minipage} 
			&  
			\begin{minipage}[b]{0.7\columnwidth}
				\centering
				\raisebox{-.45\height}{
					\includegraphics[width=\linewidth]{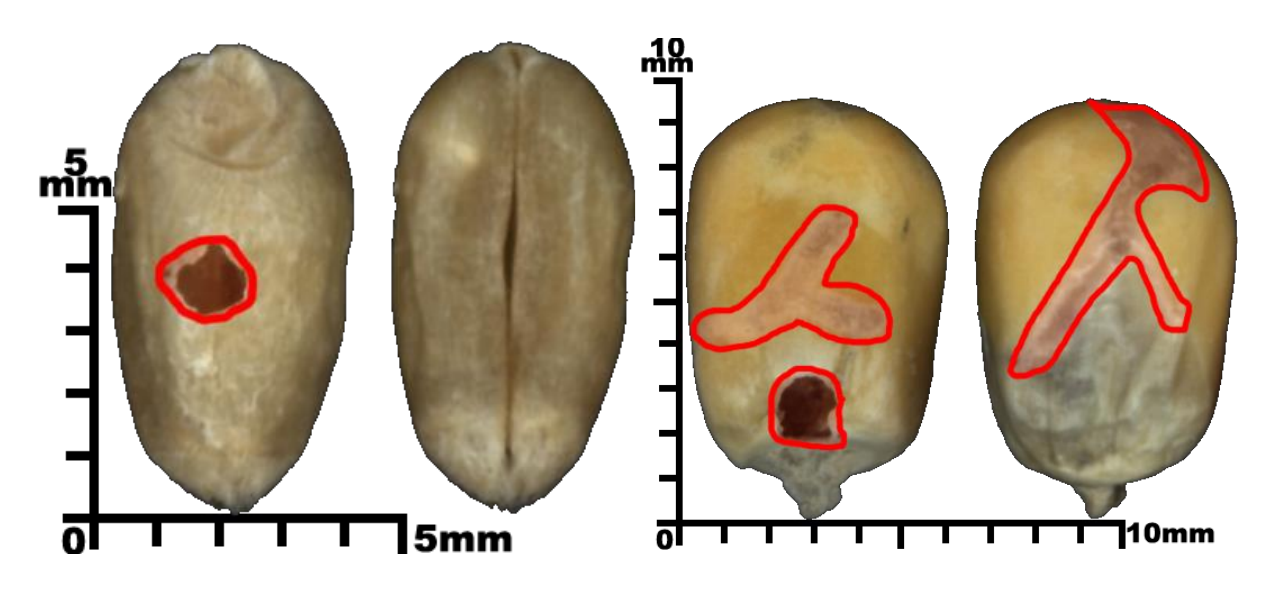}}
			\end{minipage}
			&
			\begin{minipage}[b]{0.7\columnwidth}
				\centering
				\raisebox{-.46\height}{
					\includegraphics[width=\linewidth]{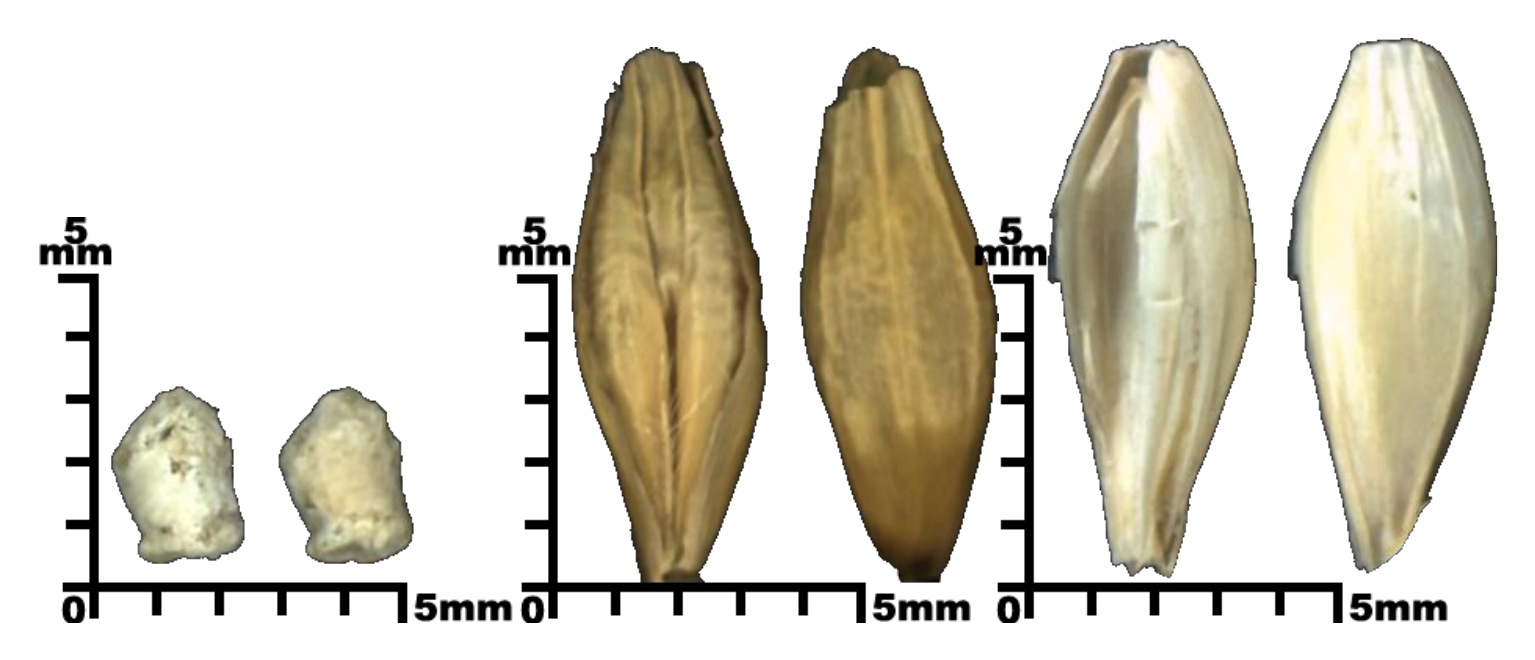}
				}
			\end{minipage} 
			 
			\\ \bottomrule
		\end{tabular}
	}
	\label{tab:wheat_category_withfig}
\end{table*}

GAI is routinely executed manually by qualified inspectors. To illustrate this process, we consider the case of inspecting a shipment of raw wheat grains (originating from granaries or freighters). According to the sampling standard \cite{ISO24333}, the procedure involves taking a laboratory sample, which amounts to 60 grams and approximately 1600 kernels. These kernels are then inspected individually in a kernel-by-kernel procedure where inspectors must carefully examine the surface of kernels and then determine them as \textit{healthy}, \textit{damaged} or other categories (see Sec.~\ref{sec:2.1}). 
However, even qualified inspectors (with 5 to 10 years of expertise) typically require around 25 minutes to complete the inspection process, since the majority of grains are small in physical size, measuring less than $8$$\times$$4$$\times4 mm^3$. As a result, inspecting these tiny grains demands a high level of concentration. Moreover, due to the nuances and superficial heterogeneity of cereal grains, manual inspection is prone to errors and lacks reliability. The available equipment or approaches for manual inspection are often cumbersome and limited in their capabilities. Therefore, in our work, we aim to develop automated GAI systems that can assist inspectors to enhance both the consistency and efficiency of inspections, providing significant social benefits.

Recently, Artificial Intelligence, particularly deep learning techniques \cite{lecun2015deep,fan2022fast}, has demonstrated an unprecedented level of proficiency, revolutionizing various fields such as medical image analysis \cite{fan2022cancer,fan2021learning}, autonomous driving \cite{driving} and anomaly detection in industries \cite{Anomaly-detection}. The widespread success of deep learning can primarily be attributed to the availability of large-scale high-quality datasets \cite{ImageNet}, sophisticated optimization objectives \cite{MSCOCO}, and advanced model architectures \cite{resnet,ViT}. The application of deep learning techniques to GAI has the potential to significantly reduce labor costs and provide more stable and efficient decision-making compared to manual inspections. We thus aim to develop an automated GAI system equipped with deep learning techniques that can have the capability to replicate the decision-making of human experts. However, the challenge that how to acquire high-quality data hampers the development of automated GAI systems. The collection of data is critical in developing robust and accurate GAI systems. The data used to train these systems must be representative of the range of samples and environments that the system will encounter in the real world, and the data must be collected and labeled with great care to ensure that it is of high quality and sufficient quantity.

In this paper, we present an automated GAI system, named \textbf{AI4GrainInsp}. It consists of data acquisition using our custom-built device, data processing for dataset creation and a deep learning-based model for GAI. Specifically, we build an automated prototype device for data acquisition (see Figure~\ref{protetypes_device}), and further annotate a large-scale dataset, called \textbf{OOD-GrainSet}, including 220$K$ single-kernel wheat or maize images with object-centric masks and corresponding healthy or damaged category information. Moreover, by integrating the cross-domain knowledge between cereal science and deep learning techniques, we formulate GAI as a ubiquitous machine learning problem, Anomaly Detection (AD) \cite{yang2021generalized}, as shown in Figure~\ref{fig:grain_examples}.b. The objective of AD is to train a model using only \textit{normal} samples, but the model is required to identify \textit{anomalous} samples during inference. For GAI, the healthy and edible kernels are considered \textit{normal} samples, while damaged kernels or other unknown objects are treated as \textit{anomalous} samples. 

We further propose an AD model for GAI, called \textbf{AD-GAI}, with a customized data augmentation strategy to synthesize anomaly-like samples based on normal samples from both image-level and feature-level perspectives. These synthesized data are used as negative samples for training a discriminator in a supervised manner. We conduct extensive experiments to verify the superior performance of AD-GAI on our \textbf{OOD-GrainSet} and the publicly available MVTec AD datasets that is typically used as the benchmark dataset for AD. \textbf{AI4GrainInsp} shows strong potential both in consistency and efficiency in comparison with human experts. The main contributions are listed as follows:
\begin{itemize}
    \vspace{-.5em}
	\setlength{\itemsep}{0pt plus 1pt}
	\item We propose an automated GAI system: \textbf{AI4GrainInsp}, which is a complete pipeline from data acquisition to deep learning-based data analysis models. 
	\item We formulate GAI as an AD problem and further propose a data augmentation-based model for GAI, called AD-GAI. Extensive experiments are conducted to verify the superiority of AD-GAI on both our grain dataset and a public benchmark dataset for AD, and validate the feasibility and efficacy of \textbf{AI4GrainInsp} both in consistency and efficiency. 
	\item We release a large-scale dataset, called \textit{OOD-GrainSet}, including $220K$ images for wheat and maize with expert-level annotations.
\end{itemize}

\section{Background}

\subsection{Grain Appearance Inspection}
\label{sec:2.1}

Wheat and maize are two of the main cereal grains and together make up approximately 42.5\% of the world's crop yield in 2022 reported in \cite{FAO_reports}. GAI serves as a requisite procedure \cite{ISO5527} for ensuring grain quality, requiring inspectors to inspect the surface of grains carefully and classify them into healthy, damaged grains, impurities and other contaminants. Damaged grains refer to grains of decreased value and can be mainly categorized into six types: sprouted (SD) grain, fusarium \& shriveled grain, black point (BP) grain for wheat or heated (HD) grain for maize, moldy (MY) grain, broken (BN) grain, grain attacked by pests (AP), as illustrated in~Table \ref{tab:wheat_category_withfig}. F\&S, MY and BP grains are contaminated by fusarium or fungus, while SD, HD, BN and AP grains have decreased values in various nutrients. On the other hand, impurities (IM), including organic objects (foreign cereals) and unknown objects (stone, plastic), can also have deleterious effects on grain processing and circulation. Similar to healthy grains, BN, AP, BP and HD are also edible to some extent. Therefore, we conduct two data partition schedules in experiments, \ie healthy grains vs. damaged grains, and edible grains vs. inedible grains.

In this paper, we propose an automated system, \textbf{AI4GrainInsp}, that utilizes a sampling device coupled with deep learning techniques. Considering the heterogeneity and diversity of grains, we formulate GAI as an anomaly detection problem, and demonstrate our \textbf{AI4GrainInsp} equipped with deep learning techniques increases the accuracy and efficiency of the inspection process.

\subsection{AI for Smart Agriculture and Food}

In recent years, artificial intelligence (AI) techniques have achieved significant progress in the field of smart agriculture \cite{mitra2022everything}. For example, by analyzing satellite images or drone images, AI can forecast and monitor climatic and soil conditions, as well as predict crop yield and production \cite{eli2019applications}. AI-based Unmanned Aerial Vehicles (UAV) and autonomous tractors have provided robust navigation and dynamic planning techniques for smart irrigation and disease control \cite{virk2020smart}. With the help of remote cameras, AI is also capable of analyzing plant diseases and detecting pest distributions \cite{cardim2020automatic}. Furthermore, some researchers have attempted to use AI to recognize food categories \cite{Food-Recognition}, and estimate the calorie and nutrition content \cite{Nutrition5k}.

However, there has been limited research \cite{fan2022grainspace} on cereal grains in the cultivation-grain-processed food streamline. Grain quality determination still lags behind, with no automated devices currently available and manual-inspection strategies proving to be cumbersome. In this paper, we focus on this critical yet underestimated field of grain quality determination, especially GAI. We demonstrate that building an automated GAI system is a highly challenging problem. We endeavor to build an effective system powered by deep learning techniques, to ensure food safety and contribute to the development of smart agriculture and promoting progress toward ``Good Health and Well-being'' and Sustainable Development Goals.

\subsection{Anomaly Detection}

Visual anomaly detection \cite{yang2021generalized,Liu_2018_CVPR,georgescu2021anomaly} means that only normal samples are available during training time, while normal and anomalous samples should be identified during inference. Early studies attempt to formulate anomaly detection as one-class classification~\cite{ruff2018deep,perera2019ocgan,liznerski2020explainable} that assign high confidence to \textit{in-distribution} samples and low probabilities to \textit{out-of-distribution} samples, and there is a line of work called SVDD-based methods~\cite{tax2004support,ruff2018deep} that train models to project representations into a hypersphere space. 

The majority of recent deep learning-based studies adopt reconstruction-based methods. These methods are built on a hypothesis that models can effectively estimate the distribution of normal samples. These methods \cite{gong2019memorizing,zavrtanik2021draem,zavrtanik2021reconstruction,deng2022anomaly} typically adopt an encoder-decoder architecture (\eg autoencoder) to encode and decode normal images and low-dimension representations sequentially. To better learn representations, some studies \cite{gong2019memorizing,roth2022towards} introduce a memory mechanism to explicitly store different patterns of anomaly-free samples. Similar to reconstruction-based methods, some studies \cite{bergmann2020uninformed,salehi2021multiresolution} tried to learn and localize discrepancies between normal and anomalous samples by relying on knowledge distillation \cite{kd2015}. 

Recently, data augmentation-based strategy has also been widely explored \cite{li2021cutpaste,zavrtanik2021draem,liu2023simplenet,zhang2022destseg}. These methods try to synthesize anomaly-like samples based on normal samples by using well-designed data augmentation techniques, and these synthesized samples are used as supervision signals to train classification models. For example, CutPaste \cite{li2021cutpaste} employs CutMix \cite{yun2019cutmix}, DR\textit{A}EM \cite{zavrtanik2021draem} and DeSTSeg \cite{zhang2022destseg} generates anomaly-like samples based adding noise on normal images. SimpleNet \cite{liu2023simplenet} tries to identify normal features extracted from normal samples or anomalous features generated by adding noises to normal features. In this paper, our proposed AD-GAI tries to synthesize anomaly-like samples from both image-level and feature-level perspectives, achieving considerable performance on three datasets.

\begin{figure}[t]
	\centering
	\begin{center}
		\includegraphics[width=0.46\textwidth]{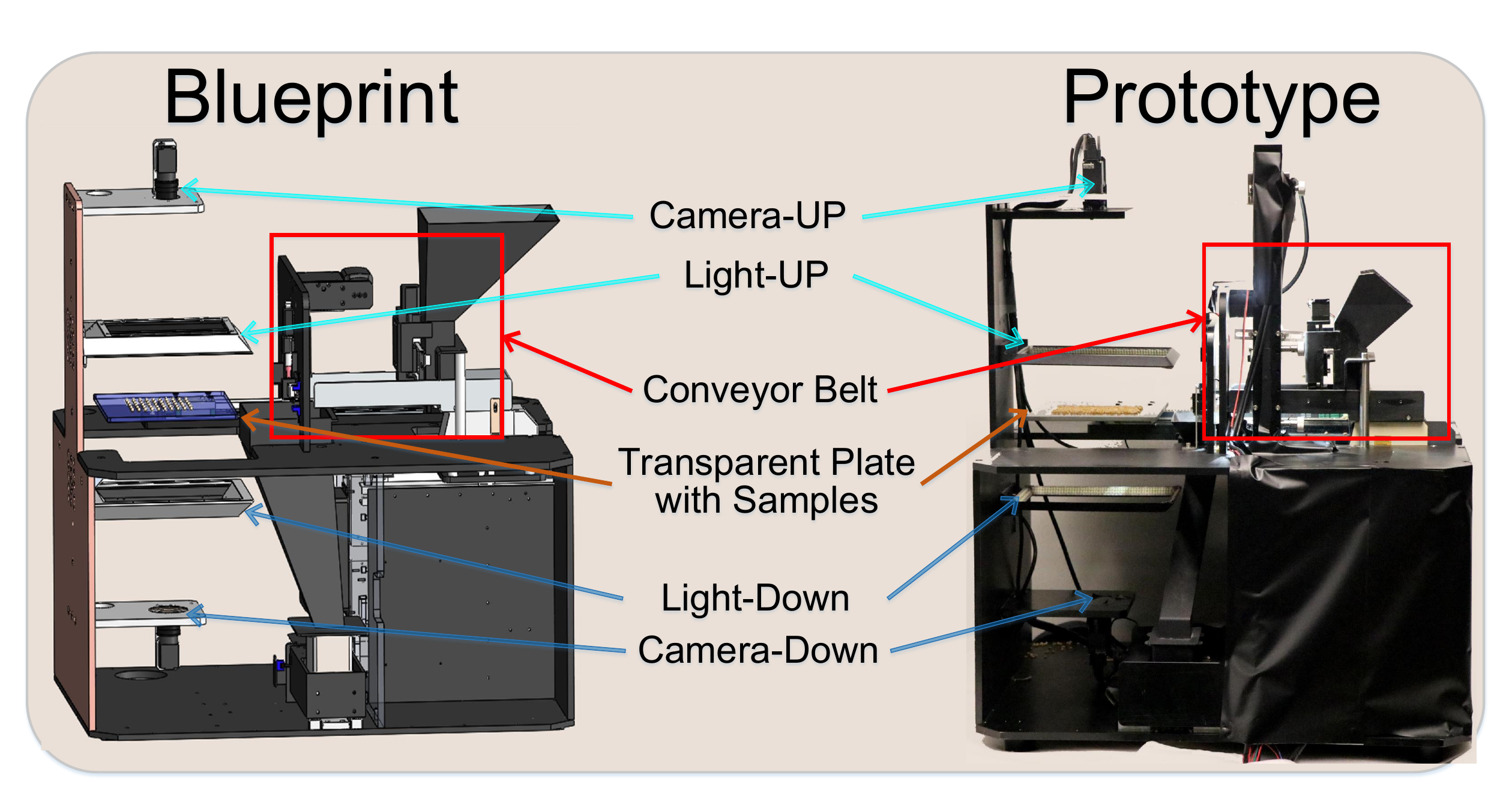}
	\end{center}
    \vspace{-1.8em}
	\caption{The blueprint and prototype device for data acquisition.}
    \vspace{-1.8em}
	\label{protetypes_device}
\end{figure}

\section{AI4GrainInsp}

\textbf{AI4GrainInsp} consists of three components: a prototype device, a large-scale dataset, \textbf{OOD-GrainSet}, and an AD grain analysis framework, \textbf{AD-GAI}. We first introduce our prototype device for data acquisition. Using this device, we captured and annotated a large-scale dataset containing about $220K$ images of single kernels with expert-level annotations. We then describe our proposed AD-GAI for automated grain quality determination.

\subsection{Data Acquisition}

There are two main challenges for capturing the visual information of raw grains: capturing high-quality images and collecting digital images efficiently. 
To overcome these challenges, we developed a customized prototype device for digitizing the surface information of grains, as shown in Figure~\ref{protetypes_device}.
For the first challenge, we employ a dual-camera strategy where two industrial cameras (860 DPI) with corresponding light sources are vertically placed along a transparent plate. We refer to the two cameras as the UP and DOWN cameras. 
To tackle the second challenge, we employ a conveyor belt with vibration bands. This enables the transparent plate to maintain a horizontal loop movement between the ends of the conveyor belt and the two cameras. The vibration bands can effectively separate the grain kernels and force kernels onto the transparent plate individually. As a result, a batch of grain kernels in the plate can be digitized together at a high sampling rate for data acquisition.

Taking a laboratory wheat sample as an example (about $60\pm 0.5$g and near 1600 kernels), to digitize the images of kernels, we divided them into several batches using the conveyor belt. Each batch consists of about 150 to 300 kernels delivered onto the transparent plate by the conveyor belt with vibration bands. Then, the plate piled with wheat kernels is placed at the center of the dual cameras. Each camera with the corresponding light source is controlled to capture high-quality images of grain kernels in a large receptive field, producing a pair of UP ($I_{up}$) and DOWN ($I_{down}$) images from the two cameras for a batch of kernels. Finally, we obtain several pairs of high-quality images for a laboratory wheat sample.

\subsection{OOD-GrainSet}
 
\begin{figure}[tb]
	\begin{center}
		\includegraphics[width=0.48\textwidth]{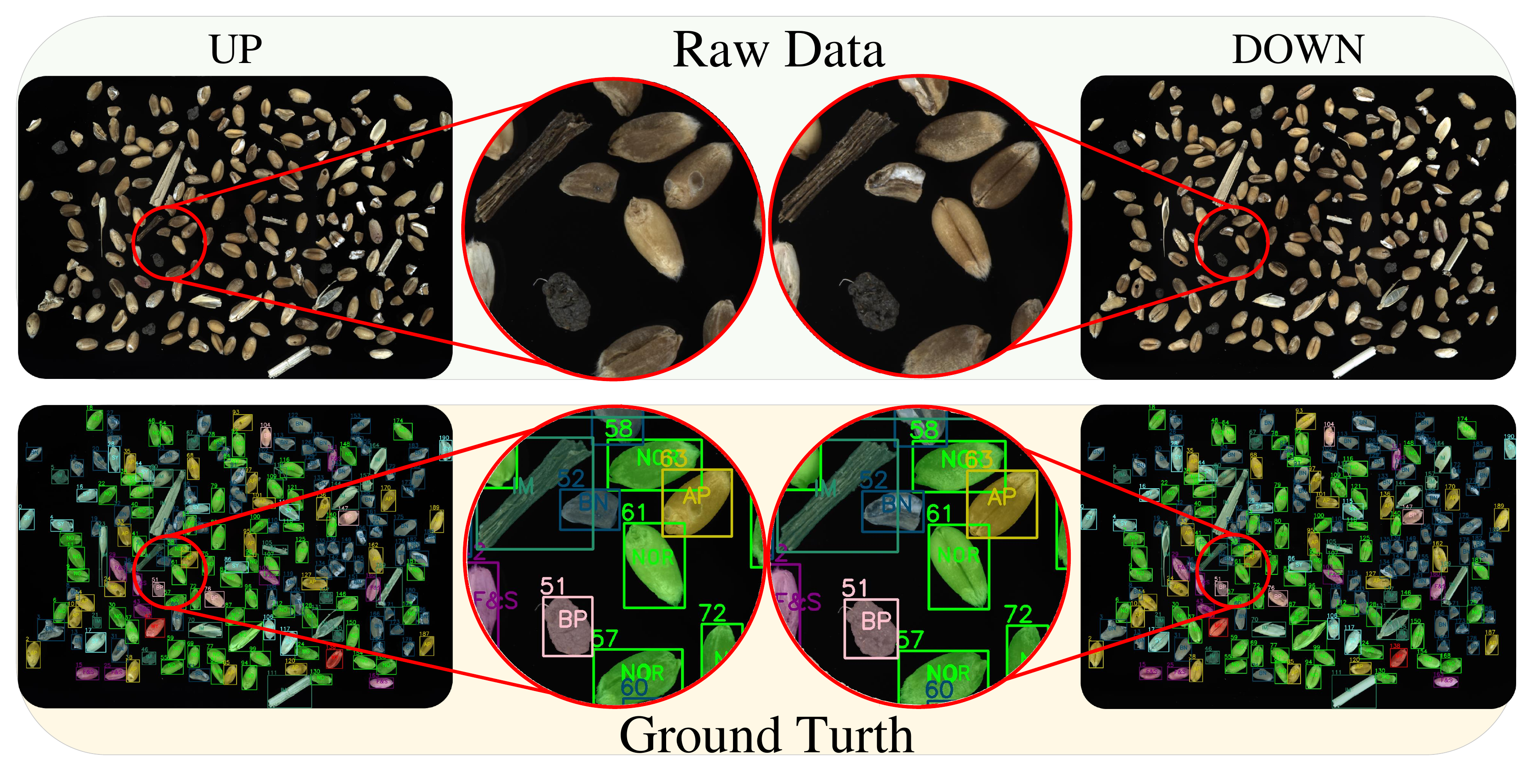}
	\end{center}
    \vspace{-1em}
	\caption{Annotation examples for a batch of grain kernels. Both high-resolution UP ($I_{up}$) and DOWN ($I_{down}$) images contain a set of grain kernels. The pair information, object localization and morphological shape are healthy or damaged grains (denoted in different colors) are provided.}
	\label{fig:data}
    \vspace{-.5em}
\end{figure}

\textbf{Raw Data:}
Figure~\ref{fig:data} illustrates an example of a pair of images captured from UP ($I_{up}$) and DOWN ($I_{down}$) angles, each of which has a high resolution of $3644$$\times$$5480$ pixels covering $ 91$$\times$$137mm^2$. As $I_{up}$ and $I_{down}$ are totally vertical to the transparent plate, the combination of UP and DOWN images covers about $92$ to $98\%$ of the superficial areas of grain kernels according to physical measurements.

\textbf{Expert-level Annotations:}
As a pair of UP and DOWN images ($I_{up}$ and $I_{down}$) capture the surface information from the top and bottom views, each kernel in these images has two sides and shares the same healthy or damaged grain category information. Inspired by rotation object detection \cite{xia2018dota} and instance segmentation \cite{MSCOCO} tasks, we annotate these images from four perspectives: single-kernel pair information, object localization, kernel mask and damaged grain category (see Figure~\ref{fig:data}). For example, a pair of UP and DOWN images produce a set of single-kernel images containing two sides of kernels in a horizontal layer, where each image has a corresponding segmentation mask $M$ depicting the morphological shape at the pixel-level. All single kernels are classified as healthy, impurities, or one of the six damaged grain categories. To simplify the processing for building the AD dataset, all single-kernel images are processed with geometrical transformations to show similar poses, as shown in Table~\ref{tab:wheat_category_withfig}.

\begin{figure}[t]
	\begin{center}
		\includegraphics[width=0.46\textwidth]{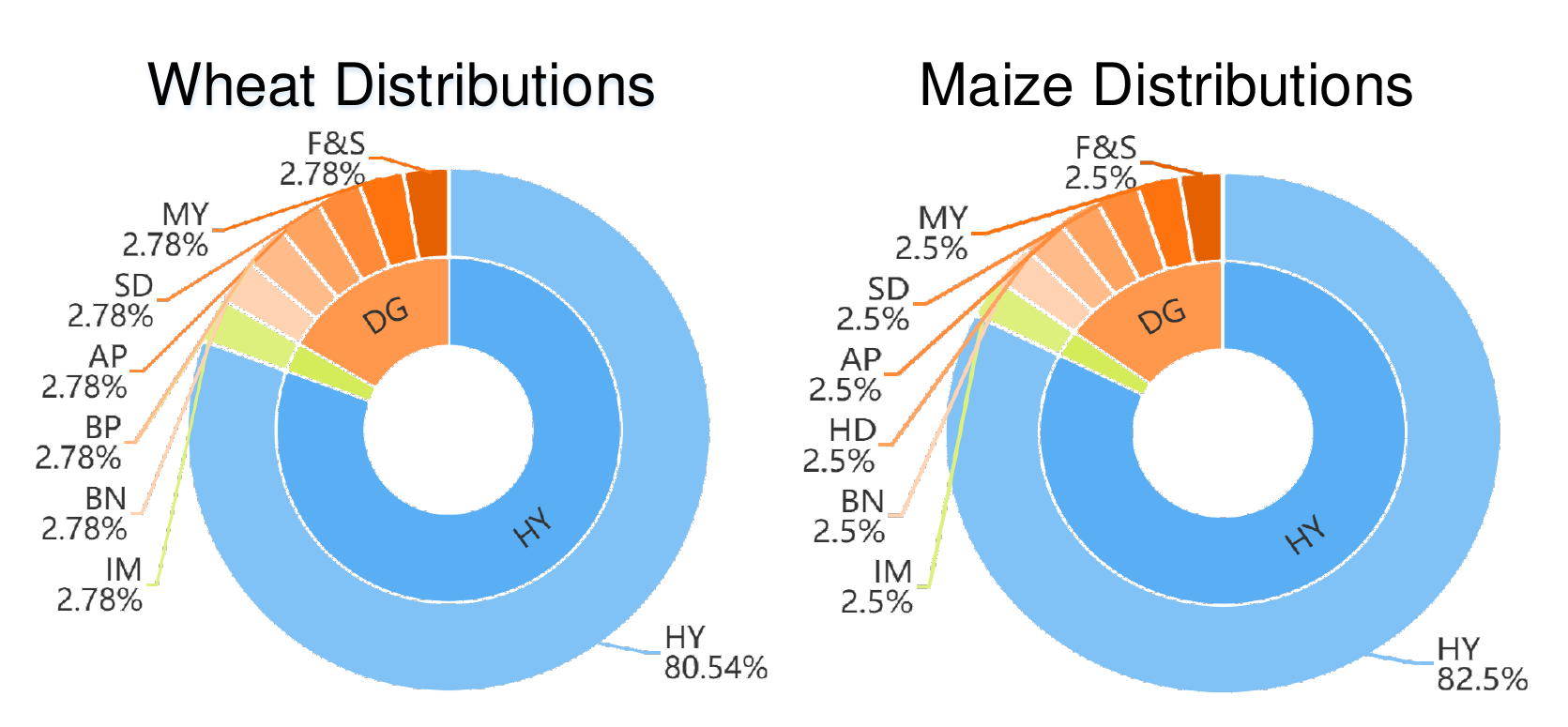}
	\end{center}
    \vspace{-1.5em}
	\caption{The distributions of \textit{OOD-GrainSet}, including healthy, damaged grains (DG) and impurities for wheat and maize. Among damaged grains, the categories of BN, AP, and BP/HD are classified as edible.}
    \vspace{-1em}
	\label{fig:data-dist}
\end{figure}

\textbf{Distributions of OOD-GrainSet}. Our dataset\footnote{More details can be found on the project website.}, called \textbf{\textit{OOD-GrainSet}}, involves two types of cereal grains: wheat and maize. Given the nature of grains, the proportion of damaged grains is relatively small and we made efforts to maintain a balanced distribution for building OOD-GrainSet, as shown in Figure~\ref{fig:data-dist}. For wheat data, we annotated about $180K$ single-kernel images, including $145K$ healthy grains and $5K$ images for each damaged grain category and impurities. For maize data, we annotated about $40K$ single-kernel images, including $33K$ healthy grains and $1K$ images for each damaged grain category and impurities. Moreover, we additionally annotated several wheat and maize samples that are used for \textbf{AI4GrainInsp} vs. Experts experiments (see Sec.~\ref{sec:ai4grain}).

\begin{figure*}[t]
	\begin{center}
		\includegraphics[width=0.88\textwidth]{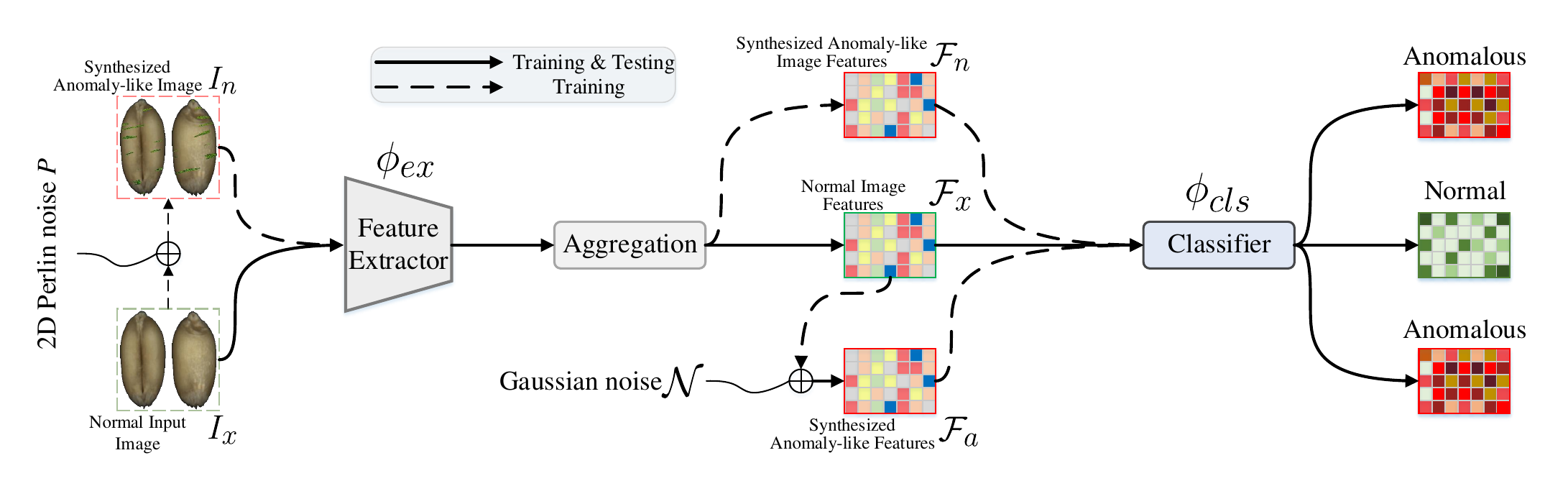}
	\end{center}
    \vspace{-1.5em}
	\caption{Overview of AD-GAI. The input normal image $I_x$ is augmented to synthesize an anomaly-like image $I_n$. Both $I_x$ and $I_n$ are fed into the feature extractor $\phi_{ex}$ to obtain patch-aware features $\mathcal{F}_x$ and $\mathcal{F}_n$ respectively. Then, $\mathcal{F}_x$ is further augmented by adding Gaussian noise to synthesize an anomaly-like feature $\mathcal{F}_a$. Finally, the classifier $\phi_{cls}$, consists of MLP layers, is trained to predict $\mathcal{F}_x$ as \textit{negative}, $\mathcal{F}_n$ and $\mathcal{F}_a$ as \textit{positive}. }
    \vspace{-1em}
	\label{fig:frameworks}
\end{figure*}

\subsection{AD-GAI}

Different from public anomaly detection data \cite{Anomaly-detection} containing rich color and contextual information collected from the wild, we consider that \textit{normal} (healthy or edible grains) and \textit{anomalous} (damaged grains or unknown objects) samples in \textbf{OOD-GrainSet} share mostly common visual information in terms of shape and context. The primary distinctions between normal and anomalous samples are fine-grained, and the characteristics of damaged grains (such as F\&S or AP) are subtle in size, such as wormholes or moldy spots. 

Based on such analysis and \textit{a priori} understanding, we propose AD-GAI which is based on a data augmentation-based strategy to synthesize anomaly-like samples. Inspired by previous AD methods for industrial inspection \cite{zavrtanik2021draem,li2021cutpaste,liu2023simplenet,zhang2022destseg}, as there are no anomalous samples that can be used during the training phase, we employ the data augmentation strategy to synthesize anomaly-like samples based on normal samples from both image-level and feature-level perspectives. These synthesized anomaly-like samples and normal samples are used together as supervision signals for training a classification model in an end-to-end manner, after which the model can identify whether test samples are normal or anomalous during inference.

\begin{algorithm}[b]
\caption{Pseudo-code of AD-GAI in a PyTorch-like style.}
\label{alg:code}
\definecolor{codeblue}{rgb}{0.25,0.5,0.5}
\lstset{
  backgroundcolor=\color{white},
  basicstyle=\fontsize{7.2pt}{7.2pt}\ttfamily\selectfont,
  columns=fullflexible,
  breaklines=true,
  captionpos=b,
  commentstyle=\fontsize{7.2pt}{7.2pt}\color{codeblue},
  keywordstyle=\fontsize{7.2pt}{7.2pt},
}
\begin{lstlisting}[language=python]
# E, C: the feature extractor and classifier
E.params = ResNet50.params  # initialize E with pre-trained ResNets
for I_x in data_loader:  # load input normal images
    I_a = aug_add_noise(I_x)  # obtain anomaly-like images: I_a
    
    F_x = E.forward(I_x)  # patch-aware features: F_x
    F_n = E.forward(I_a)  # patch-aware features: F_n
    F_a = F_x + random(N)  # obtain anomaly-like features: F_a

    loss = CrossEntropyLoss(C.forward([F_x,F_n,F_a])).mean() # channel-wise CE loss  
    loss.backward() # update C and E

\end{lstlisting}
\end{algorithm}

\textbf{Notation.} We denote $\mathbb{D}=\{ I_{0}, \dots, I_{N-1}\}$ as a training set containing only $N$ normal images and a test set $\mathbb{T}=\{ I_{0}, \dots, I_{M-1}\}$ containing $M$ normal or anomalous images. Each image $I \in \mathbb{R}^{W\times H \times C}$ ($W$, $H$ and $C$ of width, height and channels respectively) has a label $y \in \{0, 1\}$ where $0$ and $1$ mean normal or anomalous. Our goal is to train a model using only $\mathbb{D}$ during training, while the model can classify whether test samples in $\mathbb{T}$ are normal or anomalous.

The overview of AD-GAI and pseudo-code of the training procedure are presented in Figure~\ref{fig:frameworks} and Algorithm~\ref{alg:code} respectively. During the training phase, given an input image $I_{x}$ with the label $y=0$, $I_x$ is augmented by adding noise to synthesize an anomaly-like image $I_n$. Both $I_x$ and $I_n$ are fed into the feature extractor $\phi_{ex}$ to extract patch-aware features $\mathcal{F}_{x}$ and $\mathcal{F}_{n}$ respectively. Then, $\mathcal{F}_{x}$ is augmented by adding Gaussian noise to synthesize an anomaly-like feature $\mathcal{F}_{a}$. Finally, these features $\mathcal{F}_{x}$, $\mathcal{F}_{n}$ and $\mathcal{F}_{a}$ are concatenated and then fed into the classifier $\phi_{cls}$, which is optimized to discriminate these features as normal or anomalous. The details of these methods are described in the following.

\begin{figure}[t]
	\begin{center}
		\includegraphics[width=0.5\textwidth]{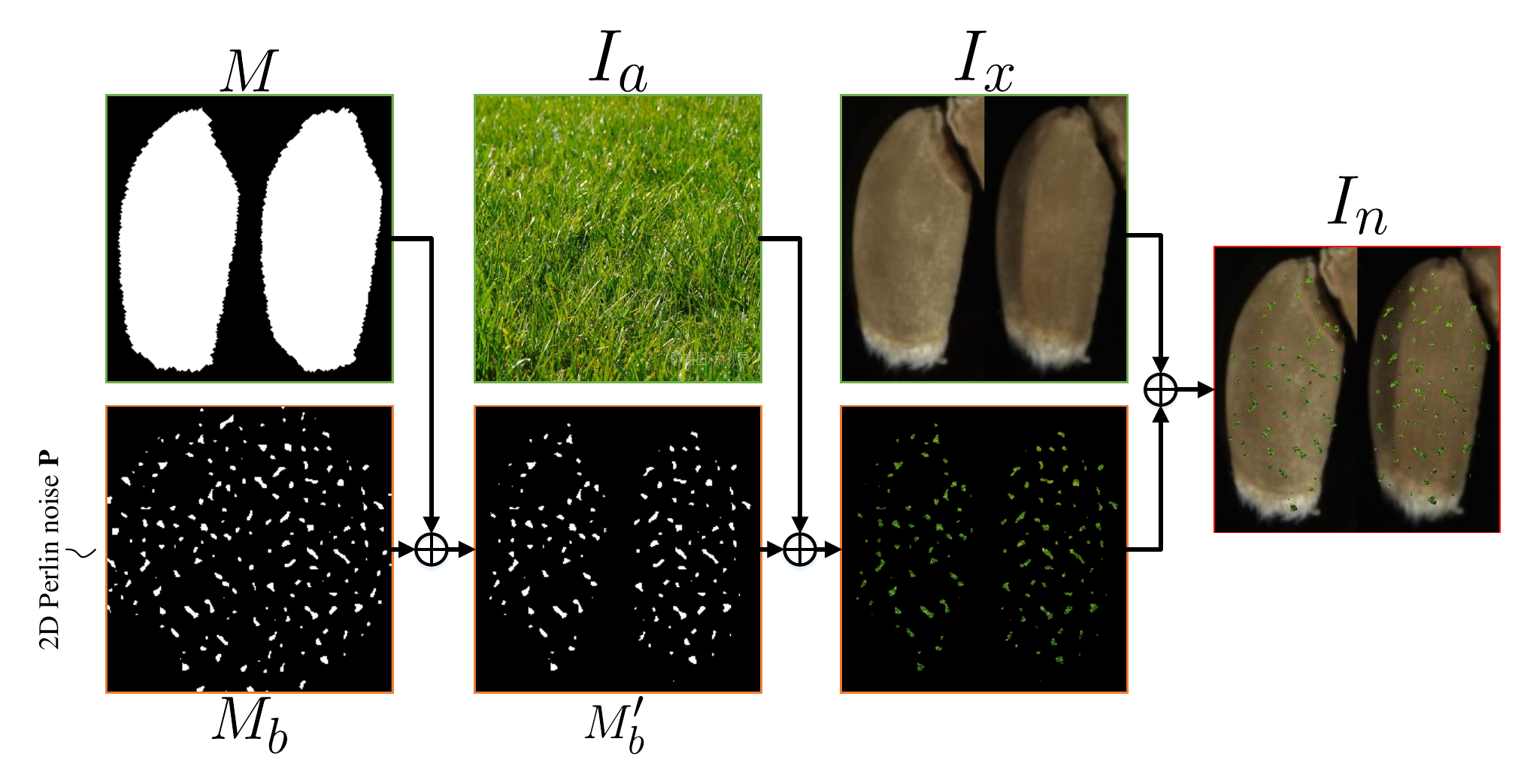}
	\end{center}
    \vspace{-1.5em}
	\caption{The simulation of image-level anomalies. $M_b$ is a binary mask used for indicating the blend between the input image $I_x$ (with a mask $M$) and an arbitrary image $I_a$ to synthesize an anomaly-like image $I_n$.}
    \vspace{-1em}
	\label{fig:gen_noise}
\end{figure}

\textbf{Simulation of image-level anomalies.} We follow the previous methods \cite{zavrtanik2021draem,zhang2022destseg} to synthesize anomaly-like image $I_{n}$ based on a normal image $I_x$, as shown in Figure~\ref{fig:gen_noise}. A binary mask $M_b \in \mathbb{R}^{W \times H}$ generated from Perlin noise $P$ contains several anomaly shapes, and an arbitrary image $I_a$ sampled from the another dataset (\eg ImageNet \cite{ImageNet}) is blended with $I_x$ based on $M_b$, which is defined as:  
\begin{equation}
    I_n = (1-M'_b)\odot I_x + \beta (M'_b \odot I_a) + (1-\beta)(M'_b \odot I_x) ,
\end{equation}
where $\beta$ is the opacity parameter between $[0.15, 1]$ as described in \cite{zavrtanik2021draem,zhang2018mixup}, and $\odot$ is the element-wise multiplication operation. $M'_b$ is generated by conducting pixel-wise \textit{and} operation between $M_b$ and the mask $M$ (provided in annotations) of the grain, which limits that generated anomaly shapes fall onto the foreground of grains.

\textbf{Extraction of patch-aware features.} We follow prior methods \cite{roth2022towards,liu2023simplenet,defard2021padim} and use a pre-trained model (\eg ResNet50 \cite{resnet} trained on ImageNet) to extract features for images. Specifically, the feature extractor $\phi_{ex}$ employs a ResNet-like model to extract hierarchical features $\{f_l \in \mathbb{R}^{w_l \times h_l \times c_l}, l\in (1,2,3,4) \}$ truncated from different convolutional stages. These features are further aggregated to obtain patch-aware features with larger receptive-of-views. For a position $(i,j)$ with the entry $f_l^{i,j}$, the aggregation is defined as:
\begin{equation}
    f'_l = \phi_{avg}( \{ f_l^{(i,j)}| (i,j) \in \mathbf{p}^* \} )  ,
\end{equation}
where $\phi_{avg}$ denotes the adaptive average pooling, and $\mathbf{p}^*$ denotes a patch centered at $(i,j)$ and a size of $p$ (set to 3). The aggregation retains the resolutions of features. To enrich feature information, we fuse aggregated features from different stages to obtain final patch-aware features. For simplification, we leverage $l=3,4$ features that contain abundant spatial and semantic information. The high-level features ($l=4$) with the smaller resolution are interpolated to the same resolution of low-level features ($l=3$), which is defined as:    
\begin{equation}
    \mathcal{F}_x = concat[f'_l, \phi_{int}(f'_{l+1})]   ,
\end{equation}
where $\phi_{int}$ and $concat$ denote the linear interpolation and channel concatenation operation respectively. The patch-aware features $\mathcal{F}_n$ can also be extracted for the synthesized anomaly images $I_n$.

\textbf{Simulation of feature-level anomalies.} Inspired by previous methods \cite{liu2023simplenet,defard2021padim}, we attempt to synthesize anomaly-like samples from the feature perspective. For the patch-aware features $\mathcal{F}_x$ extracted from the normal image $I_x$, the anomaly-like features $\mathcal{F}_a$ are synthesized by adding noise on $\mathcal{F}_x$, which is defined as:
\begin{equation}
     \mathcal{F}_a^{(i,j)} = \mathcal{F}_x^{(i,j)} + \epsilon  ,
\end{equation}
where $\epsilon$ is sampled from Gaussian distribution $\mathcal{N}(\mu,\sigma^2)$. We visualize the similarities among normal image features $\mathcal{F}_x$, anomaly-like image features $\mathcal{F}_n$ and synthesized anomaly-like features $\mathcal{F}_a$, and these features are extracted by using t-SNE technique \cite{van2008visualizing}, as shown in Figure~\ref{fig:cam_tsne}.b.

\textbf{Optimization objective.} These features are finally fed into the classifier $\phi_{cls}$. The classifier $\phi_{cls}$ employs a multi-layer perceptron (MLP) layer, and is trained to predict \textit{negative} for normal features $\mathcal{F}_x$ and \textit{positive} for anomaly-like image features $\mathcal{F}_n$ and anomaly-like features $\mathcal{F}_a$. We empirically employ cross-entropy loss (CE) as the optimization objective: 
\begin{equation}
    L = \frac{1}{w_l\cdot h_l}\sum_{(i,j)}^{w_l,h_l} CE(\phi_{cls}(\mathcal{F}^{(i,j,:)}))  .
\vspace{-.5em}
\end{equation}
where $\mathcal{F}^{(i,j,:)}$ denotes the feature vector at position $(i,j)$, and $(w_l,h_l)$ is the spatial shape of features.

\textbf{Inference}. During inference, the branches of simulation of image-level and feature-level anomalies are discarded. The output of $\max{\phi_{cls}\mathcal{F}_{I_t}^{(i,j)}|(i,j) \in \mathcal{P}^*}$ is used as the anomaly score for the test sample $I_t$ where $\mathcal{P}^*$ is the set of positions of $\mathcal{F}$.

\begin{table*}[t]
	\centering
	\caption{Comparisons of our AD-GAI and advanced methods on MVTec AD dataset and four sub-sets of our \textbf{OOD-GrainSet}.}
    \vspace{-1.5em}
	\resizebox{0.90\textwidth}{!}
	{
	\begin{tabular}{ccccccc|c}
    \toprule
    \multirow{2}[0]{*}{Methods} & \multirow{2}[0]{*}{Type}  & \multirow{2}[0]{*}{MVTec AD} & \multicolumn{5}{c}{ \textbf{OOD-GrainSet}} \\
    \cmidrule(lr){4-8}
          &          &  & Wheat(set1)  & Wheat(set2) & Maize(set1)  & Maize(set2) & Average \\
    \cmidrule(lr){1-8}
    Deep-SVDD (ICML-18 \cite{ruff2018deep}) & Distance-based  & 59.2 & 86.5  & 86.6 & 80.9  & 76.4  & 82.6 \\
    PADiM (ICPR-21 \cite{defard2021padim}) & Distance-based  & 95.8 & 73.1  & 67.5 & 67.2  & 59.4  & 66.8 \\
    Mem-AE (ICCV-19 \cite{gong2019memorizing}) & Reconstruction-based  & -  & 85.8  & 84.9 & 73.8  & 56.4  & 75.2 \\
    DR\textit{A}EM (ICCV-21 \cite{zavrtanik2021draem})  & Reconstruction-based & 98.1 & 79.8  & 59.5 & 66.4  & 78.1  & 70.9 \\
    RevDist (CVPR-22 \cite{deng2022anomaly}) & Reconstruction-based & 98.4  & 90.1  & 89.2 & 86.5  & 81.9  & 86.9 \\
    CSI  (NeurIPS-20 \cite{tack2020csi}) & Data Augmentation-based  & -  & 83.6  & 77.3 & 84.7  & 78.6  & 81.1 \\
    CutPaste (CVPR-21 \cite{li2021cutpaste}) & Data Augmentation-based   & 96.1 & 76.7  & 77.5  & 75.1  & 71.3  & 75.2\\
    \cmidrule(lr){1-8}
    AD-GAI (single R50 model) &  Data Augmentation-based   & 99.0 & 94.2  & 93.5 & 87.7
  & 82.5  & 89.5 \\
      AD-GAI (ensemble R50\&R101) &  Data Augmentation-based   & \underline{\textbf{99.1}} & \underline{\textbf{95.9}}  & \underline{\textbf{94.1}} & \underline{\textbf{88.2}}
  & \underline{\textbf{82.8}}  & \underline{\textbf{90.2}} \\
    \bottomrule
    \end{tabular}%
	}
	\label{tab:sota}
\end{table*}

\section{Experiments}

\subsection{Experimental Settings}

\textbf{Datasets.} We explore our AD-GAI on our OOD-GrainSet and the public benchmark MVTec AD \cite{Anomaly-detection}. The \textbf{MVTec AD dataset} \cite{Anomaly-detection} is widely used for evaluating anomaly detection methods. It provides 5354 high-resolution images across 10 object categories and 5 texture categories, such as toothbrush and wood. The training set comprises 3629 normal images, while the test set includes 1725 normal or anomalous images along with pixel-level anomaly annotations. We follow the experimental settings as \cite{zavrtanik2021draem,deng2022anomaly} where we train an individual model for each category. 

For our \textbf{OOD-GrainSet}, we construct four sub-sets according to the grains' conditions, as shown in Table~\ref{tab:data-set}. Wheat(set1) and Maize(set1) indicate that only healthy grains are treated as \textit{normal} samples, while the remaining grains are considered anomalies. Wheat(set2) and Maize(set2) mean that healthy grains and some of the damaged yet edible grains are combined as \textit{normal} samples. We split normal samples into 70\% and 30\% partitions for the training and test sets, preserving the ratio of different categories. All anomalous samples are used for the test set.

\textbf{Implementation details.} All experiments are conducted on a workstation with RTX 3090 GPUs based on the PyTorch platform \cite{paszke2019pytorch}. We use ResNet-50 \cite{resnet} pre-trained on ImageNet \cite{ImageNet} as the default feature extractor. We employ Adam optimizer \cite{kingma2014adam} with momentum of $(0.8, 0.999)$, weight decay of $1 \times 10^{-4}$, initial learning rate of $1\times 10^{-3}$. The batch size is set to 4 and the training epoch is set to 8 and 16 for wheat and maize respectively.

\textbf{Evaluation metrics}. we use the commonly used Area Under
the Receiver Operating Curve (AUROC) as the metric for both MVTec AD and OOD-GrainSet. In addition, to validate \textbf{AI4GrainInsp} with human experts, we employ the Macro F1-score (threshold is set to 0.3 for our AD-GAI) as the metric, and we also report the inspection time to evaluate the runtime efficiency.

\begin{table}
	\centering
    \vspace{-1em}
	\caption{Detailed settings of two subsets: \textbf{set1} and \textbf{set2}. $\checkmark$ and $\bigcirc$ indicate that the category is used as \textit{normal} or \textit{anomalous} samples. Only healthy grains are treated as \textit{normal} samples in the \textbf{set1}, while edible grains are treated as \textit{normal} samples in the \textbf{set2}.}
    \vspace{-1em}
    \resizebox{0.48\textwidth}{!}
	{
    \begin{tabular}{ccccccccc}
    \toprule
    \multirow{2}[0]{*}{\makecell[c]{Dataset \\ Settings}} &  \multirow{2}[0]{*}{HY}      & \multicolumn{6}{c}{Damaged Grains}            & \multirow{2}[0]{*}{IM}  \\
    \cmidrule(lr){3-8}
          &     & BN    & AP    & BP/HD & SD    & FS    & MY    &  \\
    \cmidrule(lr){1-9}
    Wheat/Maize(set1) & $\checkmark$     &  $\bigcirc$     & $\bigcirc$     & $\bigcirc$     & $\bigcirc$    & $\bigcirc$     & $\bigcirc$     & $\bigcirc$ \\
    Wheat/Maize(set2) & $\checkmark$    & $\checkmark$     & $\checkmark$     & $\checkmark$     & $\bigcirc$     & $\bigcirc$     & $\bigcirc$     & $\bigcirc$ \\
    \bottomrule
    \end{tabular}%
	}
    \vspace{-1.6em}
	\label{tab:data-set}
\end{table}

\subsection{Comparisons with Advanced Methods}

The experiments were conducted on MVTec AD and four sub-sets of OOD-GrainSet, by comparing with three types of AD methods: distance-based (Deep-SVDD \cite{ruff2018deep} and PADiM \cite{defard2021padim}), reconstruction-based (Mem-AE \cite{gong2019memorizing}, DR\textit{A}EM \cite{zavrtanik2021draem} and RevDist \cite{deng2022anomaly}) and data augmentation-based methods (CSI \cite{tack2020csi} and CutPaste \cite{li2021cutpaste}). 

As shown in Table~\ref{tab:sota}, our \textbf{AD-GAI} produces the best performance on all four sub-sets of OOD-GrainSet, achieving about 5.8\%, 4.9\%, 1.7\% and 0.9\% improvement over other advanced methods on Wheat(set1), Wheat(set2), Maize(set1) and Maize(set2) respectively. Moreover, our AD-GAI also produces excellent results on the MVTec AD dataset, with 99.1\% of image-level AUROC performance. Compared to other data augmentation-based methods, our model achieves substantial improvements, which validates the effectiveness of our AD-GAI that attempts to synthesize anomaly-like samples from both image-level and feature-level perspectives.

\subsection{Ablation Study}

\textbf{Backbones for feature extractors}. The feature extractor $\phi_{ex}$ extracts patch-aware features from input images. We test different ResNet-like \cite{resnet} backbones without data augmentation, as shown in Table~\ref{tab:abl-bk_aug}. We observe that using R50 pre-trained on ImageNet gains significant improvements of 14.4\% and 19.5\% on Wheat(set1) and Maize(set1) compared to R50 from scratch, which confirms the effectiveness of using pre-trained models to extract features. We further explore backbones with different parameter scales. Using lightweight model R18 pre-trained on ImageNet also outperforms R50 from scratch, and larger models R50 and R101 can produce better performance than R18. It is noted that R50 and R101 show similar performance, and we select R50 as our default backbone due to its relatively lower computational costs.

\begin{table}
	\centering
    \vspace{-1em}
	\caption{Ablation study of different backbones and data augmentation techniques on Wheat(set1)/Maize(set1).}
    \vspace{-1em}
    \resizebox{0.48\textwidth}{!}
	{
    \begin{tabular}{ccccc}
    \toprule
    \multirow{2}[0]{*}{Backbone} & R50 from scratch & R18 & R50 & R101 \\
          & 79.8/67.2  & 88.9/80.7  & \textbf{94.2}/\textbf{87.7}  & 94.1/86.3 \\
    \midrule
    \multirow{2}[0]{*}{\makecell[c]{Data \\ Augmentation}}  & None  & Flip+Rot  & Mixup \cite{zhang2018mixup} & RandAug \cite{cubuk2020randaugment} \\
          & \textbf{94.2}/\textbf{87.7}  & 90.3/87.4  & 93.5/87.1  & 89.7/83.2 \\
    \bottomrule
    \end{tabular}%
	}
	\label{tab:abl-bk_aug}
\end{table}

\textbf{Data augmentations}. We conducted experiments by using different data augmentation techniques. Compared to Flip+Rot (horizontal and vertical flipping and 90, 180, 270 rotations), mixup \cite{zhang2018mixup} or RandAug \cite{cubuk2020randaugment}, it is noted that using no data augmentation (in addition to our sample synthesis) produces the best performance of 94.2\% and 87.7\% on Wheat(set1) and Maize(set1) respectively. We consider it is because both training and test samples are well-processed during data annotations, and using heavy data augmentation techniques can be harmful to simulations of anomaly-like samples since the distinctions between normal and anomalous samples are subtle.  

\begin{figure}[ht]
	\centering
	\begin{center}
		\includegraphics[width=0.34\textwidth]{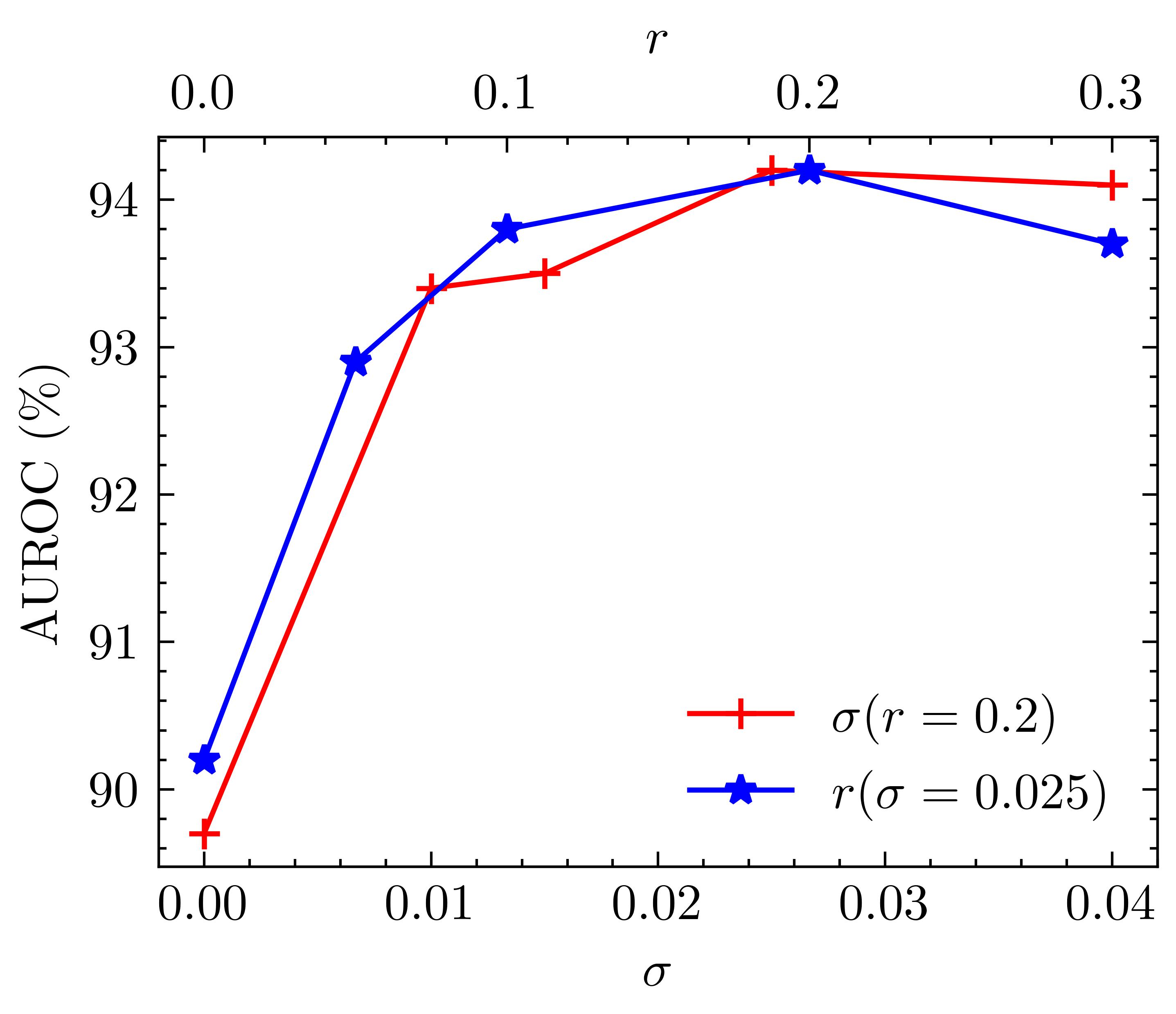}
	\end{center}
    \vspace{-2.5em}
	\caption{The ablation study of noise levels on Wheat(set1). $r$ and $\sigma$ can enable or control noise levels of image-level and feature-level simulations.}
    \vspace{-1em}
	\label{fig:noise-level}
\end{figure}

\textbf{Structure of AD-GAI and noise levels}. We also conducted experiments on Wheat(set1) to investigate the impact of noise levels on both image-level and feature-level simulations of anomalies, as shown in Figure~\ref{fig:noise-level}. We formulate a noise parameter $r$ that represents the ratio of the maximum area of binary noise mask $M_b$ to the mask of grain $M$. Particularly, using only image-level (\ie $\sigma=0$) or feature-level (\ie $r=0$) simulations produces moderate results of 89.7\% and 90.2\% respectively, which confirms the effectiveness of using both simulations together. AD-GAI achieves the best performance of 94.2\% when $\sigma=0.025$ and $r = 0.2$. We consider that small values of $\sigma$ or $r$ cannot synthesize anomalies well, while large values will produce redundant anomaly-like samples harmful to training an effective classifier with limited normal samples.

\begin{figure}[!h]
	\centering
    \vspace{-1em}
	\begin{center}
		\includegraphics[width=0.46\textwidth]{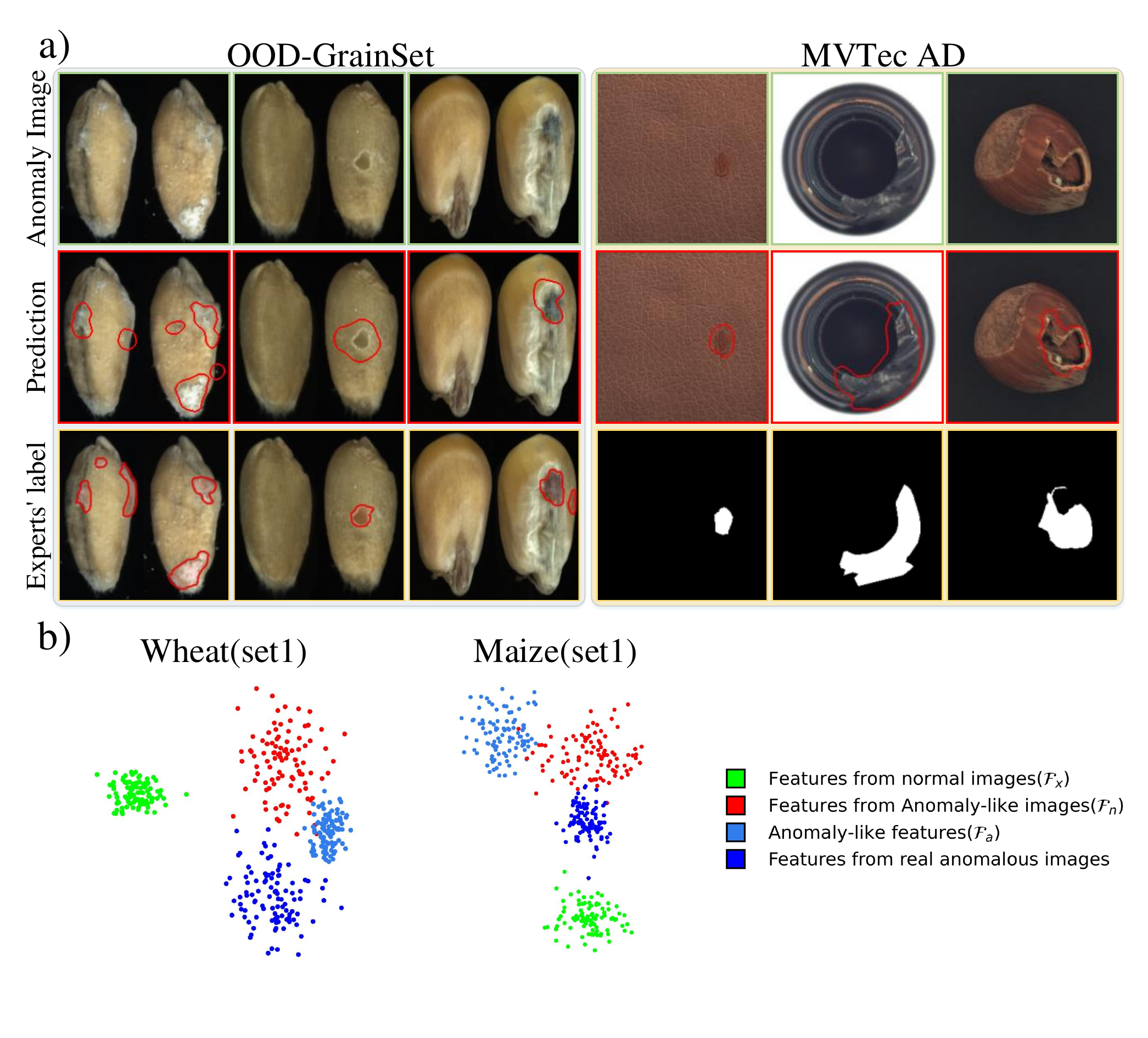}
	\end{center}
    \vspace{-3em}
	\caption{a) Visualization of anomaly images, prediction of AD-GAI with Grad-CAM technique \cite{selvaraju2017grad} and experts' annotations. b) Visualization of features from normal, anomalous and simulations by using t-SNE \cite{van2008visualizing}.}
    \vspace{-3em}
	\label{fig:cam_tsne}
\end{figure}

\textbf{Qualitative analysis}. We utilize the Grad-CAM technique \cite{selvaraju2017grad} to visualize anomalous samples and prediction results from two datasets, as shown in Figure~\ref{fig:cam_tsne}.a. Our AD-GAI effectively focuses on discriminative regions, such as wormholes, moldy points, scratches, etc. Moreover, we employ the t-SNE technique \cite{van2008visualizing} to qualitatively demonstrate the similarities among features from simulations and real anomalous samples, as shown in Figure~\ref{fig:cam_tsne}.b. We observe that both image-level and feature-level synthesized anomalies are closer to real anomalous samples than normal samples, which verifies the effectiveness of our data augmentation strategies.

\subsection{\textbf{AI4GrainInsp} versus. Human Experts} 
\label{sec:ai4grain}

We further evaluate our \textbf{AI4GrainInsp} system in comparison with human experts. We enlisted two junior inspectors, JI1 and JI2, who had 3 years of experience, and two senior inspectors, SI1 and SI2, who had near 10 years of experience. We built two prototype devices, D1 and D2, each equipped with deployed AD-GAI models. We collected 4 groups of wheat samples (each of 60g) with 4\%, 8\%, 15\% and 20\% proportions of damaged grains and impurities, and 3 groups of maize samples (each of 600g, since maize grains are heavier) with 4\%, 8\% and 15\% proportions of damaged grains and impurities. We conducted and averaged two individual inspections for each test sample. We report the F1-score and time cost, which is the total running time of the system or inspectors. 

\begin{figure}[!h]
	\centering
    \vspace{-1em}
	\begin{center}
		\includegraphics[width=0.5\textwidth]{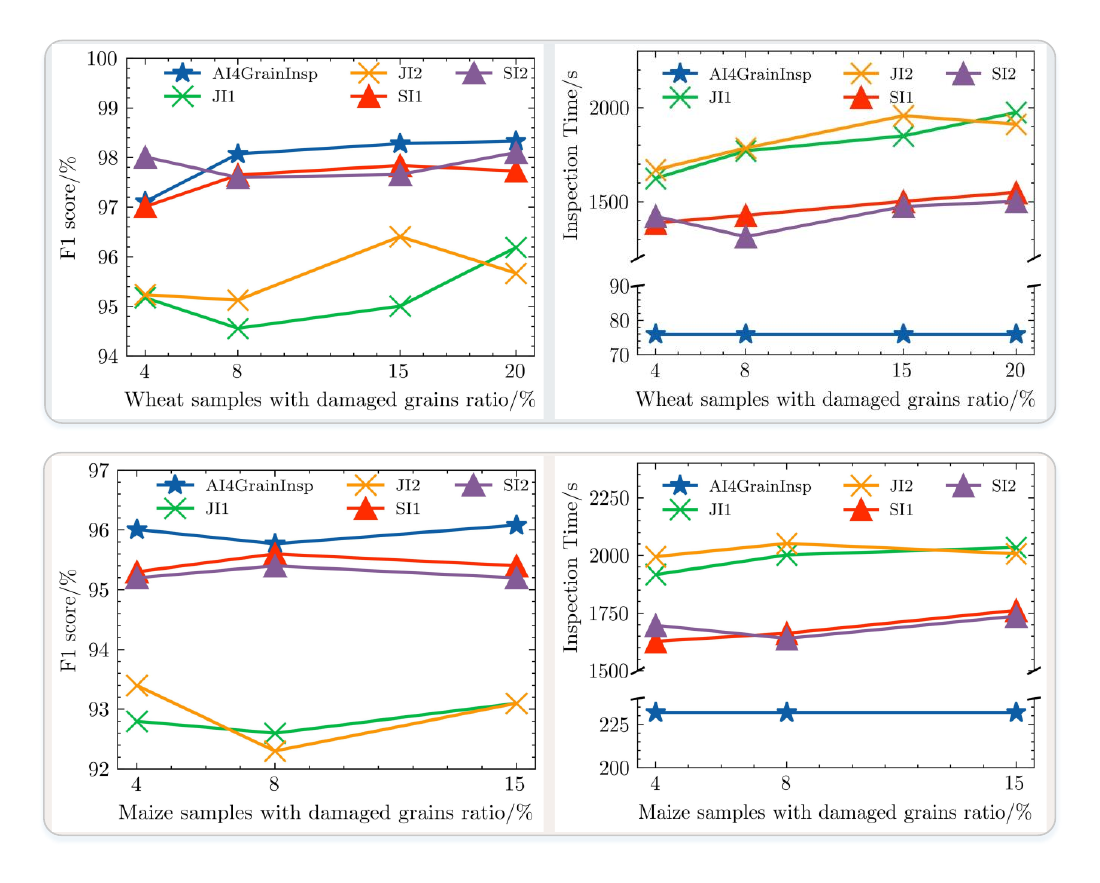}
	\end{center}
    \vspace{-2.5em}
	\caption{\textbf{AI4GrainInsp} vs. human experts on wheat and maize grains.}
    \vspace{-3.5em}
	\label{fig:ai4grain}
\end{figure}

As shown in Figure~\ref{fig:ai4grain}, our \textbf{AI4GrainInsp} produces impressive performance, which is highly consistent with senior inspectors SI1 and SI2 while being much more time-efficient over 20$\times$ speedup (about 73s vs. 1550s). Similar to wheat, experimental results on maize also validate the superiority and efficiency of our system. In contrast, the results of two junior inspectors JI1 and JI2 are relatively moderate with fluctuation, and their inspection time costs are much higher than those of devices. Therefore, we consider that our system has the potential to assist inspectors in grain quality determinations.

\section{Conclusions and Future Work}

In this paper, we propose a comprehensive automated GAI system called \textbf{AI4GrainInsp}, which includes a prototype device for data acquisition, a high-quality dataset for evaluation, and an anomaly detection model \textbf{AD-GAI}. Our model utilizes data augmentation techniques to synthesize anomaly-like samples by adding noise to normal samples from both image-level and feature-level perspectives. Experimental results demonstrate the superiority of AD-GAI, and \textbf{AI4GrainInsp} is highly consistent with human experts with much higher efficiency. Additionally, we release a large-scale dataset, \textbf{OOD-GrainSet}, containing $220K$ single-kernel images across eight categories for wheat and maize.

There still exist many challenges in GAI. For example, our \textbf{AI4GrainInsp} system coupled with a customized device has high manufacturing costs, and we aim to develop low-cost solutions, such as using smartphones to enable widespread deployment. Moreover, we also plan to train and apply \textbf{AI4GrainInsp} to more types of cereal grains, such as rice, sorghum, etc. The key challenge is to collect abundant grains from different geographical locations and build a comprehensive cereal grain atlas. We hope that our work will draw more attention to GAI-related fields and promote smart agriculture, contributing to reaching the SDG goals.

\clearpage

\bibliography{1175}
\end{document}